\documentclass[runningheads]{llncs}
\usepackage{graphicx}
\usepackage{amsmath}
\usepackage{amssymb}
\usepackage{multirow}
\usepackage{algorithm}
\usepackage{algpseudocode}  
\usepackage{color}
\usepackage{marvosym}
\usepackage{subfigure}

\usepackage{hyperref}
\hypersetup{colorlinks=true,
            linkcolor=blue,
            anchorcolor=blue,
            citecolor=blue}

\begin{document}
\title{Progtuning: Progressive Fine-tuning Framework for Transformer-based Language Models}
\titlerunning{Progtuning: Progressive Fine-tuning Framework for Transformer-based}
%
\author{Anonymous submission}
\author{Xiaoshuang Ji\inst{1,2,3} \and
Zhendong Zhao\inst{1,3}\textsuperscript{(\Letter)} \and
Xiaojun Chen\inst{1,3} \and
Xin Zhao\inst{1,2,3} \and
Zeyao Liu\inst{1,2,3}}{}
\authorrunning{X. Ji et al.}
\institute{Institute of Information Engineering,Chinese Academy of Sciences, Beijing, China \and
School of Cyber Security, University of Chinese Academy of Sciences, Beijing, China \and
Key Laboratory of Cyberspace Security Defense, Beijing, China
\email{$\{$jixiaoshuang,zhaozhendong,chenxiaojun,zhaoxin,liuzeyao$\}$@iie.ac.cn}}
%
%
\maketitle              
\begin{abstract}
Fine-tuning is a promising technique for leveraging Transformer-based language models in downstream tasks. 
As model sizes continue to grow, updating all model parameters becomes increasingly costly.
Parameter-efficient fine-tuning methods effectively address this issue by selectively updating a small subset of parameters.
However, fine-tuning and most existing parameter-efficient fine-tuning methods require updating the same number of parameters as the initial size, ignoring the unequal contribution across Transformer blocks and leading to extremely inefficient allocation of computing resources.
In this paper, we propose Progtuning, the novel fine-tuning framework combined with progressive learning for Transformer-based language models.
Specifically, Progtuning progressively reduces the number of updated transformer blocks based on the contribution.
Remarkably, Progtuning optimizes resource allocation and reduces the number of updated parameters by approximately 25\%,  while still maintaining competitive performance.
And it also exhibits high adaptability with parameter-efficient fine-tuning methods, demonstrating excellent performance across various adaptation scenarios.
\keywords{Transformer-based language model \and Parameter-efficient fine-tuning \and Progressive learning.}
\end{abstract}
\section{Introduction}
Transformer-based Language Models \cite{brown2020language,devlin2018bert,liu2019roberta,mao2024dora} have achieved great success in the realm of Natural Language Processing (NLP).
The models have been previously trained on general corpuses to learn general features, patterns, and representations.
Notable examples of Transformer-based language models include GPT-3 \cite{brown2020language}, T5 \cite{raffel2020exploring}, and LLaMA \cite{touvron2023llama}.
Fine-tuning is the predominant method to apply Transformer-based language models to down-stream tasks \cite{qiu2020pre}.
However, fine-tuning updates all parameters and stores a full copy for each task, which is gradually unacceptable as the size of models grows exponentially.
For example,  GPT-2 \cite{radford2019language} has 1.5B parameters and GPT-3 \cite{brown2020language} has 175B parameters just after one year.
Fine-tuning requires extremely expensive calculation and storage resources.
\par
To address this issue, researchers have proposed various methods, including model pruning \cite{frantar2023sparsegpt,sun2023simple}, parameter quantization \cite{dettmers2022gpt3,liu2023llm} and parameter-efficient fine-tuning \cite{han2024parameter,lialin2023scaling}.
Model pruning identifies and removes parameters or neurons that are less important, without degrading model performance.
The primary objective of pruning is to  minimize resource consumption and storage requirements, making it suitable for deployment on resource-constrained devices like mobile phones and IoT devices.
Parameter quantization reduces the precision of the numbers used to represent a model's parameters, for example, from floating-point precision (e.g., 32-bit floats) to lower precision (e.g., 8-bit integers).
The most common methods are FP8/INT8 \cite{dettmers2022gpt3} and FP4/NF4/INT4 \cite{liu2023llm}.
Parameter-efficient fine-tuning (PEFT) methods only update a small set of trainable parameters.
The trainable parameters can be a set of newly introduced modules or a subset of parameters selected from the pretrained parameters \cite{han2024parameter}.
PEFT achieves competitive performance compared with fine-tuning by updating barely 0.1\%-5\% task-specific parameters \cite{lialin2023scaling}.
At present, parameter-efficient fine-tuning is the most promising method.
\begin{figure}[h]
\vspace{-1.0em}
    \centering
    \includegraphics[width=0.6\textwidth]{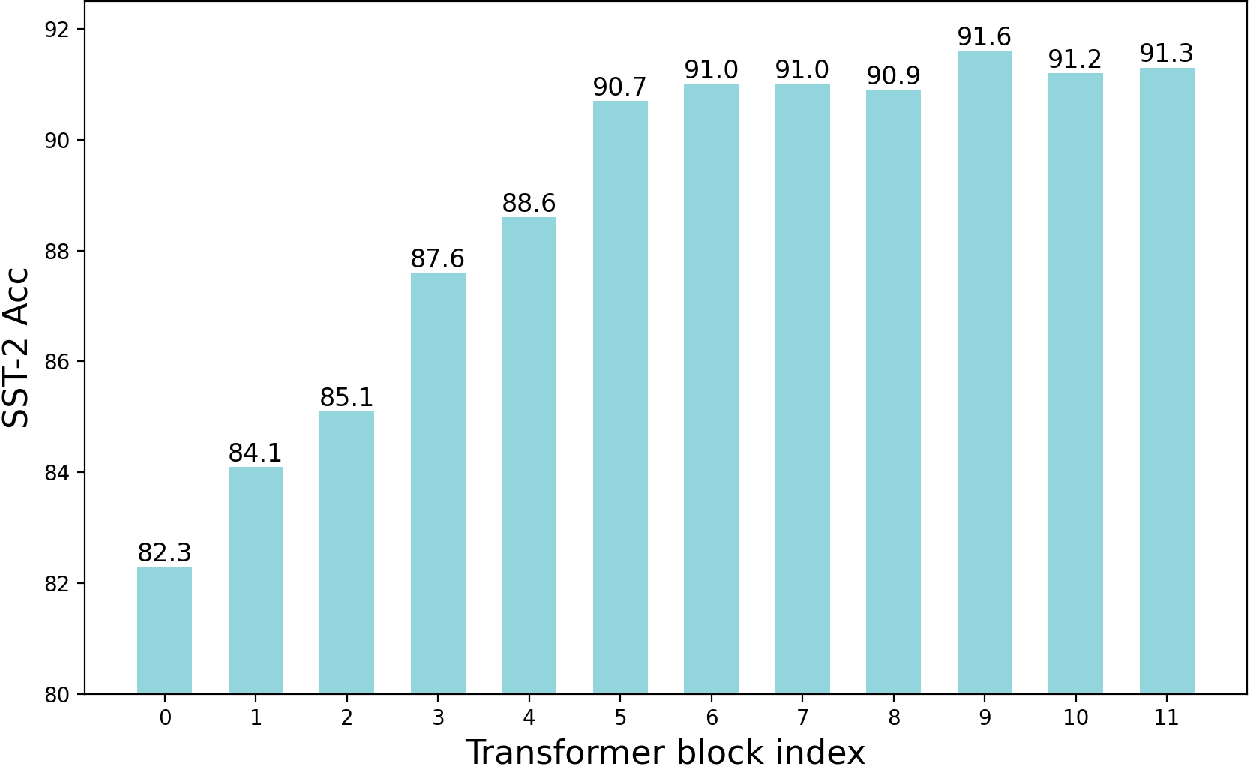}
    \caption{The performance of Transformer blocks. We only update embedding layers and selected Transformer block in fine-tuning. The model we use is $BERT_{BASE}$ and trained on SST-2 dataset.}
    \label{fig:enter-label0}
    \vspace{-1.29em}
\end{figure}
\par
However, fine-tuning and most existing PEFT methods update all trainable parameters as the initial size during the whole training process, ignoring the unequal contribution to model performance across Transformer blocks.
And this results in non-optimal allocation of computing resources.
To illustrate this point, we provide a simple example in Fig.~\ref{fig:enter-label0}.
It can be concluded that (1) the performance of Transformer blocks is different, (2) high Transformer blocks achieve better performance than low blocks.
The results motivate us to selectively allocate computing resources to high blocks that have more impact on model performance.
\par
Progtuning learning \cite{karras2017progressive} is a widely utilized technique for stabilizing the training process in the field of computer vision.
The key thought of progressive learning is to train models on simpler tasks (e.g., small images or models) and gradually extend to more complex tasks (e.g., large images or models).
And it has been widely used in super-resolution \cite{wang2018fully}, facial attribute editing \cite{wu2020cascade} and image synthesis \cite{karras2017progressive}.
In addition to stabilize the training process, computation operations are reduced inherently by progressive learning because the trained part of models is progressively changing.
\par
In this work, we propose Progtuning, the first fine-tuning framework that integrates progressive learning to gradually adjust the number of updated Transformer blocks during the training process, thereby allocating computing resource selectively.
Before the training process starts, we divide the Transformer blocks into several parts based on the number of epochs.
These parts are then organized into sequential stages, characterized by a progressive reduction in the number of Transformer blocks.
Therefore, high Transformer blocks receive more updates than low blocks during fine-tuning.
Progtuning demonstrates remarkable adaptability in combining with PEFT methods, such as Adapter tuning \cite{houlsby2019parameter}, BitFit \cite{zaken2021bitfit} and LoRA \cite{hu2021lora}.
Our experiments show that Progtuning can reduce updated parameters effectively, concurrently achieving excellent model performance.
And Progtuning seamlessly integrates with PEFT methods.
\\
Our contributions are summarized as follows.
\begin{itemize}
     \item We propose Progtuning, the first progressive fine-tuning framework.
     It gradually adjusts the amount of trainable Transformer blocks to allocate computing resources selectively.
     \item Extensive experiments are conducted on various datasets and model architectures.
     Results show that Progtuning reduces updated parameters by about 25\%.
     At the same time, the performance of models has been slightly improved.
     Progtuning also reduces training time greatly due to the reduction of updated parameters\footnote{Since the training time does not directly reflect the consumption of computing resources, we did not include it in the experiment.}.
     \item In addition, further experiments are conducted to explore the performance of combining Progtuning with PEFT methods.
     Experimental results show that the combination of Progtuning and PEFT methods works perfectly, which proves that Progtuning is of high adaptability with PEFT methods.
 \end{itemize}
\vspace{-1.0em}
\section{Related Work}
\subsection{Transformer-based Language Model}
\begin{figure}[ht]
    \centering
    \includegraphics[width=0.4\textwidth]{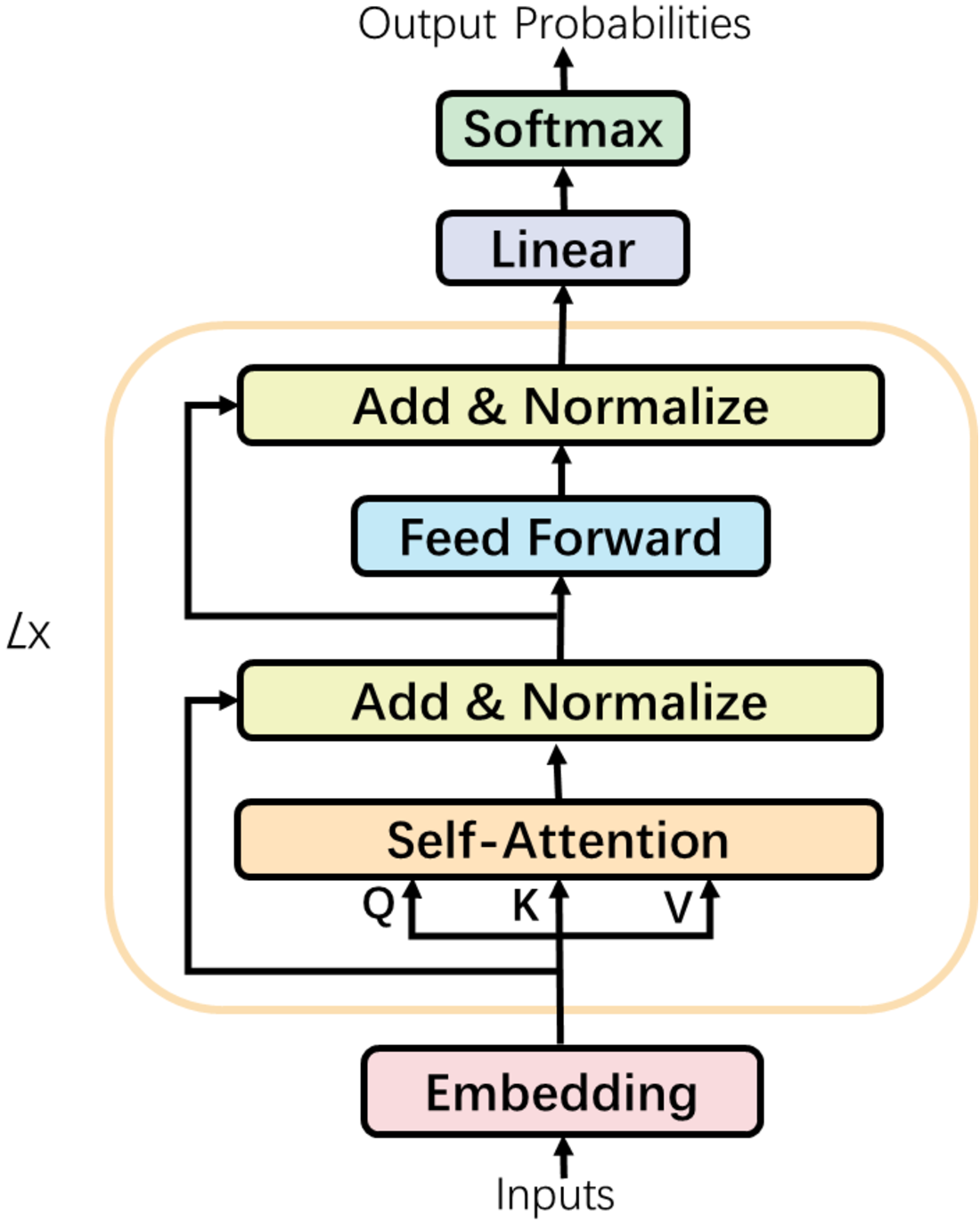}
    \caption{The structure of a Transformer-based language model.}
    \label{fig:enter-label1}
    \vspace{-1.0em}
\end{figure}

A Transformer-based language model usually consists of the embedding layers \textit{E}, several Transformers blocks and the classification header \textit{H}, as illustrated in Fig.~\ref{fig:enter-label1}.
There are stacks of \textit{L} Transformer blocks.
The embedding layers \textit{E} transform the input token lists and positional  encoding into embedding vectors and then, Transformer blocks extract semantic information individually.
Transformer block is the basic unit of pretrained language models.
A single Transformer block mainly consists of two parts: Self-Attention layer and Feed Forward Neural Network (FFN).
Both Self-Attention layer and FFN have residual connections \cite{he2016deep} and Layer Normalization \cite{ba2016layer}.
The input first passes through the Self-Attention layer to get the attention score and then is passed to the next Transformer block after being activated by FFN.
At last, the classification header \textit{H} predicts the probability distribution of the next token.
\vspace{-2.0em}
\vspace{-2.0em}
\subsection{Fine-tuning}
Fine-tuning is the predominant model training technique used in the realm of NLP \cite{dai2015semi,howard2018universal}.
Pretrained models are trained on specific datasets to acquire domain-specific knowledge of downstream tasks.
Fine-tuning modifies all parameters and saves a full copy of pretrained parameters for each task.
Since the model has been pretrained on the general corpuses, the number of fine-tuning epochs is usually small but significantly enhances the model performance.
The emergence of BERT \cite{devlin2018bert} has broken the record held by the traditional language models.
With the rapid development in deep learning and hardware, numerous large pretrained models have emerged, such as GPT-3 \cite{brown2020language}, T5 \cite{raffel2020exploring} and LLaMA \cite{touvron2023llama}. 
The amounts of pretrained model parameters rapidly grow to hundreds of billions, which makes updating all parameters unacceptable.
\vspace{-1.0em}
\subsection{Parameter-efficient Fine-tuning}
Although fine-tuning has been successful in many tasks, it still has several limitations.
For example, fine-tuning updates all parameters of models, which is unacceptable due to the explosive growth of model parameters.
Recent years have witnessed the rise of parameter-efficient fine-tuning methods, known as PEFT.
PEFT freezes most parameters and updates a small set of parameters to reduce computing resource consumption while doing no harm to model performance.
There are three mainstream classes of methods: addition-based, selection-based, and reparametrization-based \cite{lialin2023scaling}.
The addition-based PEFT methods freeze pretrained model parameters and add new parameters or modules, such as Adapter tuning \cite{houlsby2019parameter}, Prefix-tuning \cite{li2021prefix} and Prompt tuning \cite{lester2021power}.
The selection-based PEFT methods select a subset of parameters and freeze the others.
Typical methods include BitFit \cite{zaken2021bitfit} and FAR \cite{vucetic2022efficient}.
The reparametrization-based methods introduce reparametrization to reduce trainable parameters, such as LoRA \cite{hu2021lora} and KronA \cite{edalati2022krona}.
Building upon these methods, many variant methods have emerged, such as P-tuning \cite{liu2023gpt}, P-tuning V2 \cite{liu2021p}, QLoRA \cite{dettmers2024qlora} and LoftQ \cite{li2023loftq}.
\section{Method}
\begin{figure}[htb]
    \vspace{-2.0em}
    \centering
    \includegraphics[width=1\textwidth]{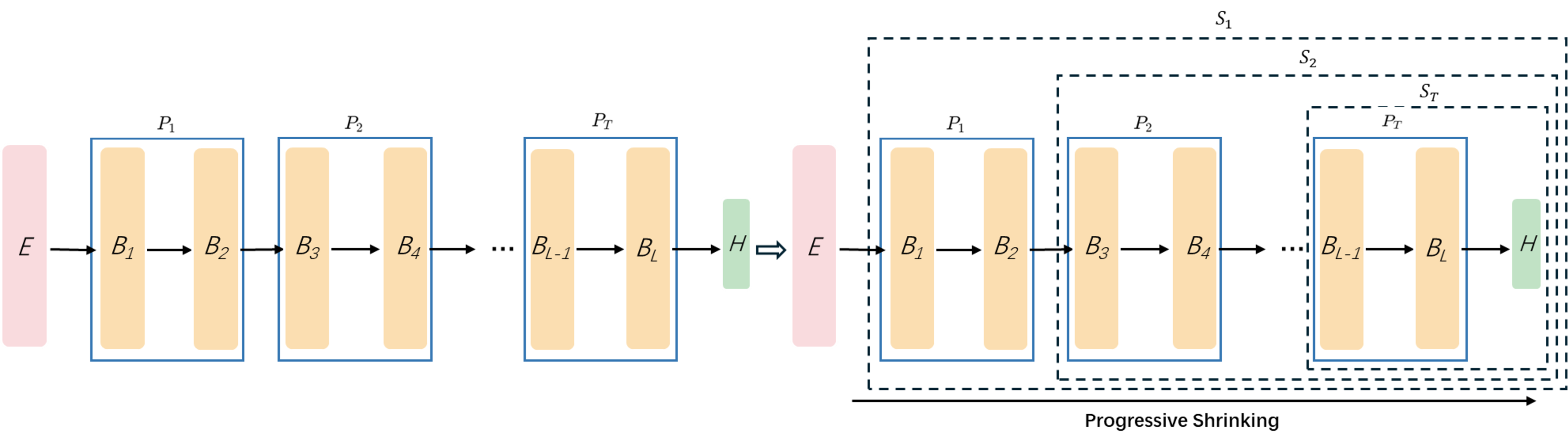}
    \caption{The overview of Progtuning. We divide the model into \textit{T} parts and arrange them to form overlapping and shrinking stages.}
    \label{fig:enter-label2}
    \vspace{-1.5em}
\end{figure}

For a given Transformer-based language model $\mathcal{M}$, it consists of the embedding layers \textit{E}, several Transformers blocks and the classification header \textit{H}.
The number of Transformer blocks is denoted by \textit{L} and numbered from low to high layers, namely $B_1$, $B_2$ ... $B_L$.
The model $\mathcal{M}$ can be expressed as:
\vspace{-0.6em}
\begin{equation}
    \mathcal{M}:=\textit{E}\circ\operatorname*{\sum}_{i=1}^{\textit{L}}B_{i}\circ H=\textit{E}\circ B_{1}\circ B_{2}\circ\cdots \circ B_{\textit{L}} \circ H.
    \vspace{-0.5em}
    \label{eq:model}
\end{equation}
To combine with progressive learning, the model is first divided into several parts .
Suppose that the number of fine-tuning epochs is \textit{T}, for $\textit{T} \in \mathbb{Z}^+$.
We can divide Transformer blocks into \textit{T} parts, denoted by $P_t$, for $ t \in \{1,...,\textit{T}\}$ and number them sequentially.
Each part comprises $\lfloor \frac{\textit{L}}{\textit{T}} \rfloor$ Transformer blocks as shown in Fig.~\ref{fig:enter-label2} (left).
Consequently, the model is composed of several parts, the header \textit{H} and the embedding layers \textit{E}, namely,
\vspace{-0.3em}
\begin{equation}
    \mathcal{M}:=\textit{E}\circ\operatorname*{\sum}_{i=1}^{\textit{T}}P_{i}\circ H=\textit{E}\circ P_{1}\circ P_{2}\circ\cdots \circ P_{\textit{T}} \circ H.
    \vspace{-0.8em}
    \label{eq:part}
\end{equation}
\\
Then we arrange these parts to construct \textit{T} overlapping and shrinking stages, denoted by $\textit{S}_t$, for $ t \in \{1,...,\textit{T}\}$.
Overlapping means Transformer blocks are shared by all stages.
Shrinking means the number of Transformer blocks is decreasing.
The stage $S_t$ is the combination of the parts numbered from t to \textit{T}, along with the classification header \textit{H} as shown in Fig.~\ref{fig:enter-label2} (right).
Formally, we define
\begin{equation}
    S_t:=\operatorname*{\sum}_{i=t}^{\textit{T}}P_{i}\circ\textit{H}= P_{t}\circ P_{t+1}\circ \cdots\circ P_{\textit{T}}\circ\textit{H}.
    \label{eq:stage}
\end{equation}
\vspace{-0.8em}
\\
During fine-tuning, we just need to train the i-th stage and the embedding layers \textit{E} in the i-th epoch, for i $\in \{1,...,\textit{T}\}$.
This ensures that only the parameters of the i-th stage get modified while the others keep frozen in the i-th epoch.
As the training proceeds, the capacity of trainable stages decreases from the initial stages $S_1$ = $\mathcal{M}$ to the final stages $S_\textit{T}$.
In this way, computing resources get allocated selectively to high Transformer blocks.
\par
Moreover, Progtuning seamlessly integrates with PEFT methods by replacing the whole Transformer blocks with the trainable parameters.
For example, if we combine Progtuning with Adapter tuning, the adapter modules rather than the whole Transformer blocks are divided and updated.
The entire process is summarized in Algorithm~\ref{alg:Framwork}.
\vspace{-1.0em}
\begin{algorithm}[htb] 
  \caption{Progtuning---Fine-tuning Framework with Progressive Learning}  
  \label{alg:Framwork}  
  \begin{algorithmic}[1]
    \Require  
        fine-tuning epochs: \textit{T}; model: $\mathcal{M}$; \textit{fine-tuning type} choice between fine-tuning and Progtuning; \textit{PEFT method} choice between Adapter tuning, BitFit and LoRA.
     \Ensure 
     \If{$algorithm$ == Adapter tuning or BitFit or LoRA }
        \State add or select trainable parameters and freeze others
     \EndIf
     \If{$\textit{fine-tuning type}$ == Progtuning}
        \State divide trainable parameters into \textit{T} parts as Eq.~\ref{eq:part}
        \State arrange divided parts to form \textit{T} stages as Eq.~\ref{eq:stage}
         \For{i=1 to \textit{T}}
            \State update parameters of i-th stages
        \EndFor
    \Else
    \For{i=1 to \textit{T}}
        \State update all trainable parameters
        \EndFor
     \EndIf
  \end{algorithmic}
\end{algorithm}
\vspace{-3.0em}
\section{Experiments and Results}
\subsection{Experiments Settings}
\textbf{Datasets and Metrics} We evaluate Progtuning on the GLUE benchmark \cite{wang2018glue} and SQuAD \cite{rajpurkar2016squad} datasets.
The GLUE benchmark is a collection of nine natural language understanding tasks, including single-sentence classification tasks, similarity and paraphrase tasks, and inference tasks.
Consistent with prior research \cite{guo2020parameter,houlsby2019parameter,hu2021lora,zaken2021bitfit}, we exclude the WNLI task, where BERT does not outperform other models. 
There are two versions of SQuAD, which are SQuAD v1.1 and SQuAD v2.0.
SQuAD v1.1 is a collection of passage/question pairs and the task is to mark the answer text span in the passage text.
SQuAD v2.0 extends SQuAD v1.1 by adding passage/question pairs that there is no answer for the question in the passage.
The metrics we use in experiments follow the settings in previous works \cite{devlin2018bert,zaken2021bitfit}.
Moreover, we track the number of updated parameters in the experiments, which are different from trainable parameters.
The number of trainable parameters counts parameters that are updatable during the training process.
The number of updated parameters is the sum of parameters that get updated in the backward propagation in each epoch.
For example, the amount of parameters of $BERT_{BASE}$ is 110M.
If we train $BERT_{BASE}$ for 3 epochs, the number of trainable parameters and updated parameters are 110M and 330M, respectively.
The number of updated parameters truly reflects the cost of computing resources and the amount of computing operation.
\\
\textbf{Models and Implementations} We conduct all experiments by using models and interfaces from the HuggingFace library \cite{wolf2020transformers}.
The models we use in experiments are $BERT_{BASE}$, $BERT_{LARGE}$ and $RoBERTa_{BASE}$. \textit{BERT} (Bidirectional Encoder Representations from Transformers) \cite{devlin2018bert} is a pretrained language model developed by Google.
It leverages a transformer architecture and is designed to understand the context of words in a sentence by considering both the left and right surroundings (bidirectional training).
\textit{RoBERTa} (Robustly optimized \textit{BERT} approach) \cite{liu2019roberta} is an advanced variant of \textit{BERT} developed by Facebook AI.
It builds upon \textit{BERT}'s architecture but introduces several key modifications to improve performance and robustness.
Both of them have several versions, differing primarily in the hidden size, the number of Transformer blocks and self-attention heads.
Base models have 12 Transformer blocks and 12 self-attention heads.
\begin{figure}[b]
    \vspace{-1.0em}
    \centering
    \includegraphics[width=0.8\textwidth]{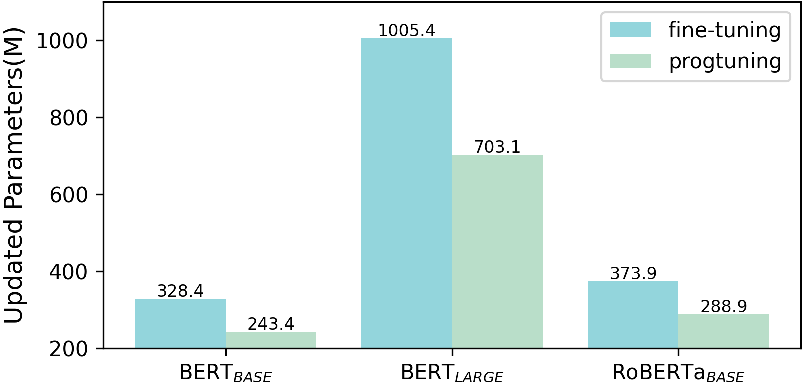}
    \caption{Updated parameters on GLUE. FT stands for fine-tuning and PT stands for Progtuning.}
    \label{fig:enter-label3}
    \vspace{-1.0em}
\end{figure}
Apart from this, the numbers of large models are 24 and 16, respectively.
As for the hidden size, it is 768 in base models and 1024 in large models.
Therefore, large models have more parameters than base models.
The code is based on \textit{Huggingface Transformers} code-base \cite{wolf2019huggingface}, chosen for its cleanliness and well-structured design.
Every task is performed 5 times on NVIDIA-A100 GPUs to get the average result.
\vspace{-1.0em}
\subsection{Results of Progtuning}
\textbf{GLUE} Fig.~\ref{fig:enter-label3} and Table~\ref{tab1} show the results of Progtuning and fine-tuning on the GLUE benchmark after 3 epochs.
Fig.~\ref{fig:enter-label3} shows the number of updated parameters on GLUE benchmark.
Generally, Progtuning reduces updated parameters by approximately 25\%.
The updated parameters of the $BERT_{LARGE}$ model is even reduced by 30\% from 1005.4M to 703.1M.
Table~\ref{tab1} shows the performance of models.
It can be seen that Progtuning can not only reduce the number of updated parameters, but also enhance the model performance.
Notably, the average score of $RoBERTa_{BASE}$ is even 0.7 point higher than fine-tuning.
We analyze that the enhancement is attributed to Progtuning's approach of progressively adapting the number of updated Transformer blocks, thereby mitigating overfitting that can occur from repeatedly updating all model parameters.
Therefore, Progtuning yields superior results on the test set.
We find that $BERT_{LARGE}$ is sometimes unstable on some tasks, which is consistent with the original paper \cite{devlin2018bert}.
So, we only use $BERT_{BASE}$ to conduct follow-up experiments.
\vspace{-1.0em}
\begin{table*}[htb]
\caption{Model performance on GLUE. FT stands for fine-tuning and PT stands for Progtuning. The best results are shown in \textbf{bold}.}
\label{tab1}
\scalebox{0.85}{
\begin{tabular}{c|c|ccccccccc|c}
\hline
                      Model & Method & CoLA & $MNLI_m$ & $MNLI_{mm}$ & MRPC & QNLI & QQP  & RTE  & SST-2 & STS-B & Avg \\ \hline
\multirow{2}{*}{$BERT_{BASE}$} & FT & 58.2 & \textbf{84.6}  & 83.7   & \textbf{89.9} & \textbf{91.4} & \textbf{88.2} & 66.5 & \textbf{92.7}  & \textbf{88.4}  & 82.6 \\  
                      & PT & \textbf{59.6} & 83.9  & \textbf{84.0}   & 89.8 & 91.0 & 87.6 & \textbf{68.7} & 92.0  & 88.3  & \textbf{82.8} \\ \hline
\multirow{2}{*}{$BERT_{LARGE}$} & FT & 58.7 & 85.8 & \textbf{86.0} & \textbf{89.9} & \textbf{92.3} & \textbf{88.2} & 67.7 & 93.2 & 89.2 & 83.4 \\  
                      & PT & \textbf{59.6}    & \textbf{86.1} & \textbf{86.0} & 89.5 & 91.9 & 88.0 & \textbf{70.3} & \textbf{93.5} & \textbf{89.7}      & \textbf{83.8} \\ \hline
\multirow{2}{*}{$RoBERTa_{BASE}$} & FT & 58.1 & 85.5 & 85.7 & \textbf{92.1} & 91.3 & 86.5 & 63.2 & 93.1 & \textbf{89.8} & 82.8 \\  
                      & PT & \textbf{59.0} & \textbf{86.3} & \textbf{86.4} & 91.8 & \textbf{91.9} & \textbf{87.2} & \textbf{65.8} & \textbf{93.3} & 89.6 & \textbf{83.5} \\ \hline
\end{tabular}
}
\end{table*}
\vspace{-1.0em}
\\
\textbf{SQuAD} Table~\ref{tab2} presents the results on SQuAD datasets.
The numbers of epochs on SQuAD v1.1 and SQuAD v2.0 are 3 and 2, respectively.
There are various degrees of reduction in the amount of updated parameters.
Progtuning reduces approximately 26\% and 20\% updated parameters on v1.1 and v2.0, respectively.
Apparently, the reduction of updated parameters on SQuAD v1.1 is larger because there are more epochs.
The model performance of Progtuning is better than fine-tuning whether it's on SQuAD v1.1 or SQuAD v2.0, demonstrating the competitive performance of Progtuning.
\begin{table}[htb]
\vspace{-1.0em}
\centering
\caption{Results on SQuADs. FT stands for fine-tuning and PT stands for Progtuning.}
\label{tab2}
\begin{tabular}{c|c|c|cc}
\hline
        Dataset & \begin{tabular}[c]{@{}c@{}}Method\end{tabular} & \begin{tabular}[c]{@{}c@{}}Updated\\ params (M)\end{tabular} & EM   & F1   \\ \hline
        \multirow{2}{*}{SQuAD v1.1} & FT & 326.7 & 79.0 & 87.2 \\  
        & PT & \textbf{241.6} & \textbf{80.3} & \textbf{88.1} \\ \hline
        \multirow{2}{*}{SQuAD v2.0} & FT & 217.7 & \textbf{72.5} & 74.5\\ 
        & PT & \textbf{175.3} & 71.6 & \textbf{75.0} \\ \hline
\end{tabular}
\end{table}
\vspace{-3.0em}
\subsection{Combination with PEFT Methods}
Progtuning exhibits high adaptability and works with PEFT methods perfectly.
In this section, we combine Progtuning with Adapter tuning \cite{houlsby2019parameter}, BitFit \cite{zaken2021bitfit} and LoRA \cite{hu2021lora}.
Adapter tuning introduces adapter modules into the Transformer blocks.
A adapter module consists of a down-project layer, a nonlinear layer and a up-project layer.
Specifically, the adapter modules are added after the Self-Attention layer and the FFN layer.
The thought of BitFit is quite simple, which only modifies the bias terms and the classification layers.
LoRA finds out the low-rank properties of weight matrices and introduces low-rank matrices to replace original weight matrices.
At present, LoRA is the most widely used PEFT methods.
\par
To achieve the combination with Progtuning, we replace the whole Transformer blocks with the trainable modules or parameters in their respective methods.
The adapter modules and bias terms in all Transformer block are divided in Adapter tuning and BitFit, respectively.
In LoRA, Progtuning divides the low-rank matrices rather than the whole Transformer blocks.
Experiments are conducted on GLUE benchmark and hyper parameters, such as epoch number, batch size and learning rate, are set consistent with the settings in original papers \cite{houlsby2019parameter,zaken2021bitfit,hu2021lora}.
Table~\ref{tab3} exhibits the amount of updated parameters of PEFT methods and the combination with Progtuning.
It is evident that the reduction of updated parameters is more significant on Adapter tuning and BitFit.
The reason is that the number of epochs is usually 8 or larger.
In particular, the updated parameters of Adapter tuning are even reduced by 67\%, from 35.8M to 11.9M.
Table~\ref{tab4} demonstrates that the model performance of PEFT methods and the combination with Progtuning.
Although the average results are a little lower than the original methods, the trade-off is acceptable because Progtuning helps PEFT methods reduce computing resources further.
Especially, the combination of Progtuning and Adapter tuning achieves 0.5 point lower than Adapter tuning by only updating 33\% parameters.
Considering the extremely simple algorithm and few updated parameters, the results of BitFit are acceptable.
Not surprisingly, the results of LoRA are best.
\begin{table*}[htb]
\vspace{-1.0em}
\caption{Updated parameters (M) of PEFT methods and the combination with Progtuning on GLUE.}
\label{tab3}
\scalebox{0.95}{
\begin{tabular}{c|ccccccccc}
\hline
        Method & CoLA & $MNLI_m$ & $MNLI_{mm}$ & MRPC & QNLI & QQP & RTE & SST-2 & STSB \\ \hline
        Adapter & {28.6} & {35.8} & {35.8} & {28.6} & {35.8} & {35.8} & {28.6} & {28.6} & {28.6}  \\ 
        Adapter+Progtuning & {\textbf{9.9}} & {\textbf{11.9}} & {\textbf{11.9}} & {\textbf{9.9}} & {\textbf{11.9}} & {\textbf{11.9}} & {\textbf{9.9}} & {\textbf{9.9}} & {\textbf{9.9}} \\ \hline
        BitFit & {0.8} & {1.0} & {1.0} & {0.8} & {1.0} & {1.0} & {0.8} & {0.8} & {0.8} \\  
        BitFit+Progtuning & {\textbf{0.5}} & {\textbf{0.6}} & {\textbf{0.6}} & {\textbf{0.5}} & {\textbf{0.6}} & {\textbf{0.6}} & {\textbf{0.5}} & {\textbf{0.5}} & {\textbf{0.5}} \\ \hline
        LoRA & 70.9 & 26.6 & 26.6 & 26.6 & 22.2 & 22.2 & 70.9 & 53.2 & 35.4 \\  
        LoRA+Progtuning  & \textbf{60.2} & \textbf{22.6} & \textbf{22.6} & \textbf{22.6} & \textbf{18.9} & \textbf{18.9} & \textbf{60.2} & \textbf{45.1} & \textbf{30.1} \\ \hline
\end{tabular}}
\end{table*}
\vspace{-3.0em}
\begin{table*}[htb]
\caption{Model performance of PEFT methods on GLUE. Adapter stands for Adapter tuning.}
\label{tab4}
\scalebox{0.9}{
\begin{tabular}{c|ccccccccc|c}
\hline
        Method & CoLA & $MNLI_m$ & $MNLI_{mm}$ & MRPC & QNLI & QQP & RTE & SST-2 & STSB & Avg \\ \hline
        Adapter & \textbf{60.5} & 83.8 & 84.2 & 90.8 & 91.1 & \textbf{87.4} & \textbf{72.9} & \textbf{92.8} & \textbf{89.1} & \textbf{83.6} \\
        Adapter+Progtuning & 58.8 & \textbf{83.9} & \textbf{84.3} & \textbf{90.9} & \textbf{91.2} & \textbf{87.4} & 69.6 & 92.7 & 89.0 & 83.1 \\ \hline
        BitFit & \textbf{55.9} & \textbf{78.2} & \textbf{78.7} & \textbf{90.1} & \textbf{87.5} & 82.3 & \textbf{68.8} & \textbf{91.8} & \textbf{87.2} & \textbf{80.1} \\
        BitFit+Progtuning & 54.2 & 76.8 & 77.8 & 89.8 & 87.1 & \textbf{82.4} & 68.2 & 91.5 & 86.5 & 79.4 \\ \hline
        LoRA & \textbf{64.8} & 86.4 & 85.9 & 92.5 & 92.8 & 86.2 & \textbf{79.5} & \textbf{95.0} & 90.5 & \textbf{86.0} \\
        LoRA+Progtuning & 62.9 & \textbf{86.7} & \textbf{86.2} & \textbf{92.7} & \textbf{92.9} & \textbf{86.7} & 78.8 & \textbf{95.0} & \textbf{90.6} & 85.8 \\ \hline
\end{tabular}
}
\vspace{-2.0em}
\end{table*}
\section{Ablation Experiments}
\subsection{The Necessity of Training Low Blocks}
Jawahar et al. \cite{jawahar2019does} say that the role of low Transformer blocks is to extract low-level semantic information.
Since our tasks focus more on high-level information and models we use in experiments are pretrained well, is it necessary to train low Transformer blocks? The answer is YES.
We conduct a series of ablation experiments to explore this question.
\vspace{-0.8em}
\begin{figure*}[htb]
    \centering
    \includegraphics[width=0.6\textwidth]{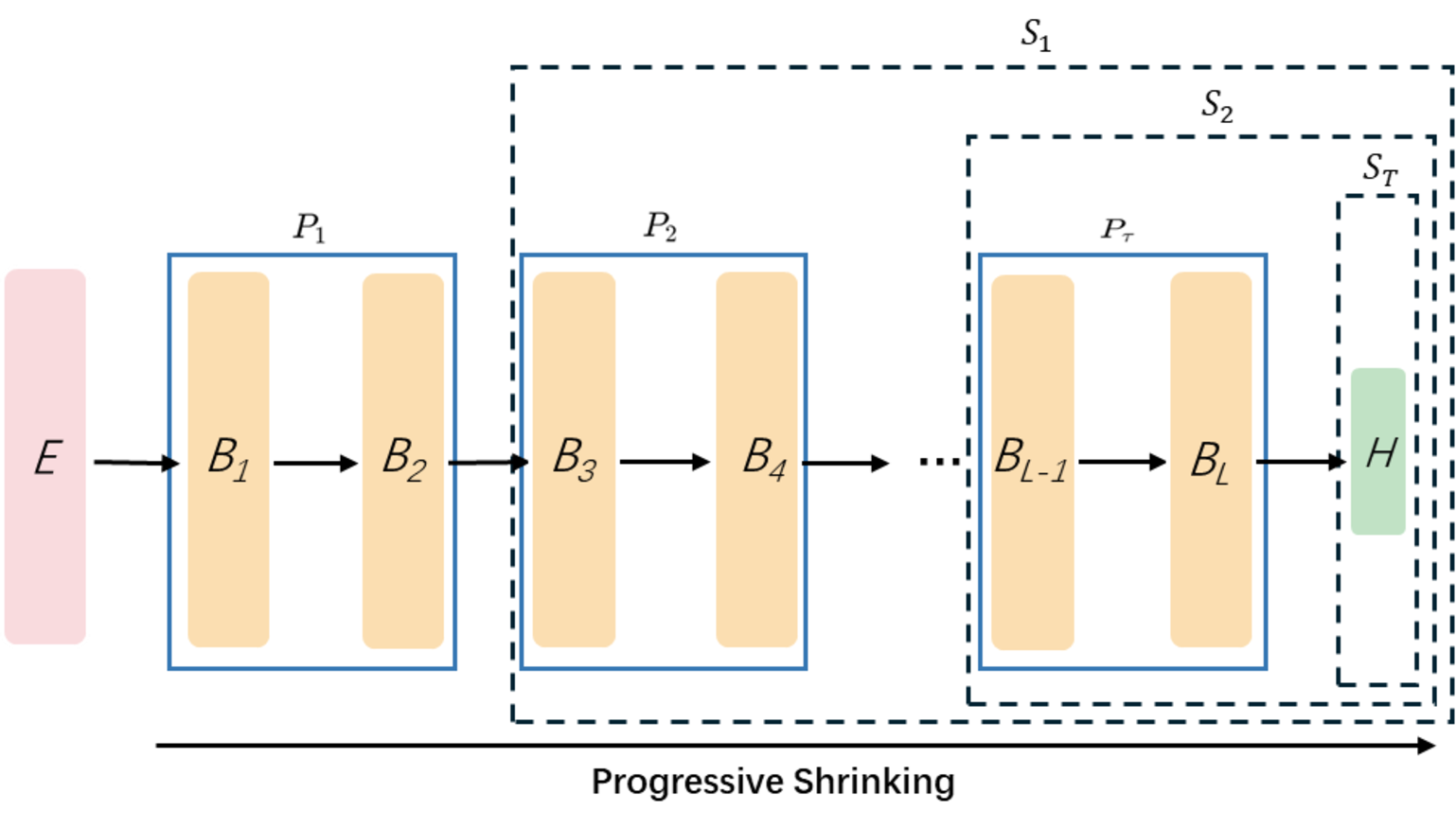}
    \caption{Progtuning without low blocks.}
    \label{fig:enter-label4}
    \vspace{-1.0em}
\end{figure*}
In experiments, we exclude the first part of every stage during training.
In other words, the first part will not be trained and the last stage only consists of the header \textit{H} as shown in Fig.~\ref{fig:enter-label4}, which can be denoted:
\begin{equation}
    S_t:=\operatorname*{\sum}_{i=t+1}^{\textit{T}}P_{i}\circ\textit{H}= P_{t+1}\circ P_{t+2}\circ\cdots\circ  P_{\textit{T}}\circ\textit{H}.
\end{equation}
\\
We use Progtuning w/o LB to represent this method.
The average results are listed in row 3 of Table~\ref{tab5}.
It is obvious that Progtuning outperforms Progtuning w/o LB on all tasks.
Pearson coefficient on the STS-B dataset is even 21.5 points lower than Progtuning.
Although the semantic information extracted by low blocks is low-level, it is still vitally important as high-level semantic information is condensed from low-level semantic information.
And low blocks need more knowledge to be an expert in the specific field.
Therefore, it is absolutely necessary training low Transformer blocks during fine-tuning.
\vspace{-1.0em}
\subsection{The Direction of Progressive Learning}
\begin{figure*}[htb]
    \centering
    \includegraphics[width=0.6\textwidth]{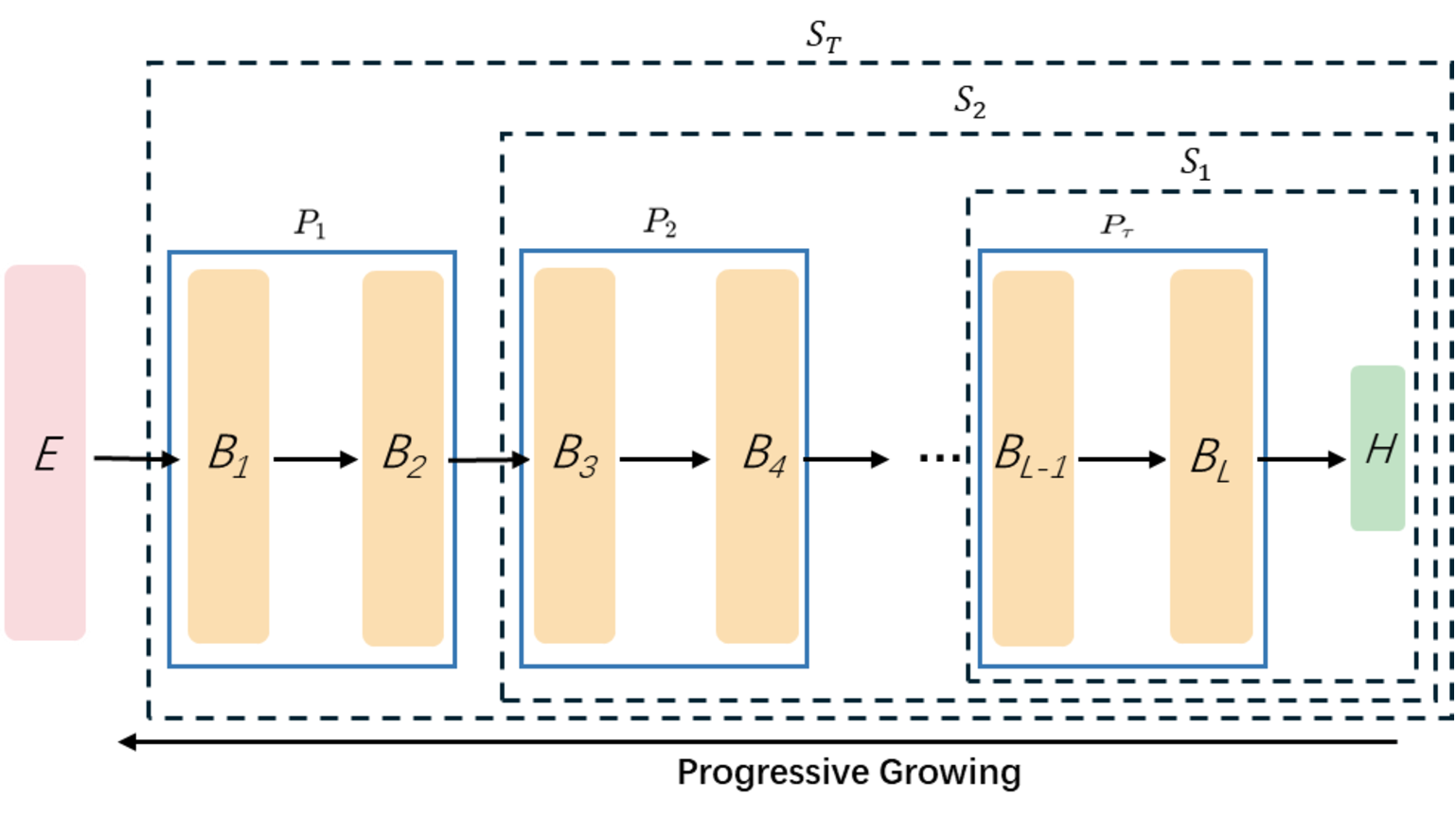}
    \caption{Progtuning that the stages are progressively growing.}
    \label{fig:enter-label5}
    \vspace{-1.0em}
\end{figure*}

There are two progressive modes, which are progressively shrinking and progressively growing.
Progressively shrinking is what we adopt in the work, in which the training procedure starts from the whole model and progressively shrinks to the last stage.
The second one is the exact opposite as shown in Fig.~\ref{fig:enter-label5}.
The first stage only contains the last part and the header \textit{H} and the subsequent stages keep growing to contain more low parts, which can be defined:
\begin{equation}
    S_t:=\operatorname*{\sum}_{i=\textit{T}-t+1}^{\textit{T}}P_{i}\circ\textit{H}=P_{\textit{T}-t+1}\circ P_{\textit{T}-t+2}\circ \cdots\circ P_{\textit{T}}\circ\textit{H}.
\end{equation}
The training procedure starts from the first stage and progressively extends to the full model, which is represented as Progtuning from HB. 
Is there any difference between the two methods?
We also conduct several ablation experiments.
The average results are shown in row 4 of Table~\ref{tab5}.
The results are lower than Progtuning and the accuracy on the MRPC dataset is even 21.9 points lower than Progtuning.
We inference that the results of Progtuning from HB are worse because low blocks can not acquire sufficient knowledge after training.
The learning rate scheduling strategy we adopt in the experiments is the linear decay strategy, in which the learning rate decreases from the preset value to zero.
So the learning rate at the last epoch is so low that the model is unable to acquire semantic knowledge.
However, the results are still a little higher than Progtuning w/o LB, which proves again that training low Transformer blocks is indispensable.
\begin{table}[htb]
\vspace{-1.0em}
\centering
\caption{Results of ablation experiments. PT w/o LB means Progtuning without training low blocks and PT from HB means Progtuning that progressively grows.}
\label{tab5}
\scalebox{1}{
\begin{tabular}{cl|cccc}
\hline
\multicolumn{2}{c|}{Method}           & CoLA & MRPC & RTE  & STS-B \\ \hline
\multicolumn{2}{c|}{Progtuning} & \textbf{59.6} & \textbf{89.8} & \textbf{68.7} & \textbf{88.3} \\ 
\multicolumn{2}{c|}{PT w/o LB}           & 54.6 & 68.5 & 65.3 & 66.8  \\ 
\multicolumn{2}{c|}{PT from HB}           & 56.5 & 67.9 & 67.6 & 68.5  \\ \hline
\end{tabular}}
\end{table}
\vspace{-2.0em}
\section{Conclusion}
We propose Progtuning, the first framework that integrates progressive learning with fine-tuning.
The thoughts of Progtuning are simple and effective, which gradually adjusts the number of Transformer blocks involved in the backward propagation to allocate computing resources selectively.
Experiment results show the superior performance of Progtuning on various tasks and models.
Progtuning can reduce updated parameters by 25\% compared with fine-tuning.
At the same time, the model performance is even better than fine-tuning attributed to the reduction of over-fitting.
Progtuning also exhibits high adaptability in working with parameter-efficient fine-tuning methods to reduce computing resources further.
\section*{Acknowledgement}
This work is supported by Beijing Municipal Science and Technology Commission (No. Z231100005923047).
%
%
%
\bibliographystyle{splncs04}
\bibliography{0023}
%




\end{document}